\documentclass[letterpaper, 10 pt, conference]{ieeeconf}  

\IEEEoverridecommandlockouts                              

\usepackage{cite}
\usepackage{amsmath,amssymb,amsfonts}
\usepackage{algorithmic}
\usepackage{graphicx}
\usepackage{textcomp}
\usepackage{cuted}
\usepackage{capt-of}
\usepackage{float}
\usepackage{booktabs}
\usepackage{threeparttable}
\usepackage{mathtools}
\usepackage[mathscr]{eucal}
\usepackage[table,xcdraw]{xcolor}
\usepackage{colortbl}
\usepackage{colortbl}
\usepackage{balance}

\usepackage{algorithmic}
\usepackage{indentfirst}

\makeatletter
\let\NAT@parse\undefined
\makeatother
\usepackage{hyperref}

\title{\LARGE \bf
STEPS: Joint \underline{S}elf-supervised Night\underline{t}ime Image \underline{E}nhancement and De\underline{p}th E\underline{s}timation
}

\author{Yupeng Zheng$^{1,2}$, Chengliang Zhong$^{3,5}$, Pengfei Li$^{3}$, Huan-ang Gao$^{3}$, Yuhang Zheng$^{2}$, \\
 Bu Jin$^{1,2}$, Ling Wang$^{3,5}$, Hao Zhao$^{4}$, Guyue Zhou$^{2}$, Qichao Zhang$^{1}$ and Dongbin Zhao$^{1}$
\thanks{$^{1}$Institute of Automation, Chinese Academy of Sciences, China,
        \{zhengyupeng2022, jinbu2022, zhangqichao2014, dongbin.zhao\}@ia.ac.cn.}%
\thanks{$^{2}$Institute for AI Industry Research (AIR), Tsinghua University, China.}
\thanks{$^{3}$Department of Computer Science and Technology, Tsinghua University, China,
        \{gha20, li-pf22\}@mails.tsinghua.edu.cn.}%
\thanks{$^{4}$Intel Labs China, Peking University, China, hao.zhao@intel.com}
\thanks{$^{5}$Xi’an Research Institute of High-Tech, China.}
}

\begin{document}

\maketitle
\thispagestyle{empty}
\pagestyle{empty}

\begin{abstract}

Self-supervised depth estimation draws a lot of attention recently as it can promote the 3D sensing capabilities of self-driving vehicles. However, it intrinsically relies upon the photometric consistency assumption, which hardly holds during nighttime. Although various supervised nighttime image enhancement methods have been proposed, their generalization performance in challenging driving scenarios is not satisfactory. To this end, we propose the first method that jointly learns a nighttime image enhancer and a depth estimator, without using ground truth for either task. Our method tightly entangles two self-supervised tasks using a newly proposed uncertain pixel masking strategy. This strategy originates from the observation that nighttime images not only suffer from underexposed regions but also from overexposed regions. By fitting a bridge-shaped curve to the illumination map distribution, both regions are suppressed and two tasks are bridged naturally. We benchmark the method on two established datasets: nuScenes and RobotCar and demonstrate state-of-the-art performance on both of them. Detailed ablations also reveal the mechanism of our proposal. Last but not least, to mitigate the problem of sparse ground truth of existing datasets, we provide a new photo-realistically enhanced nighttime dataset based upon CARLA. It brings meaningful new challenges to the community. Codes, data,  and models are available at \href{https://github.com/ucaszyp/STEPS}{https://github.com/ucaszyp/STEPS}.

\end{abstract}
\section{Introduction}
\begin{figure}
  \centering
  \includegraphics[width=0.45\textwidth]{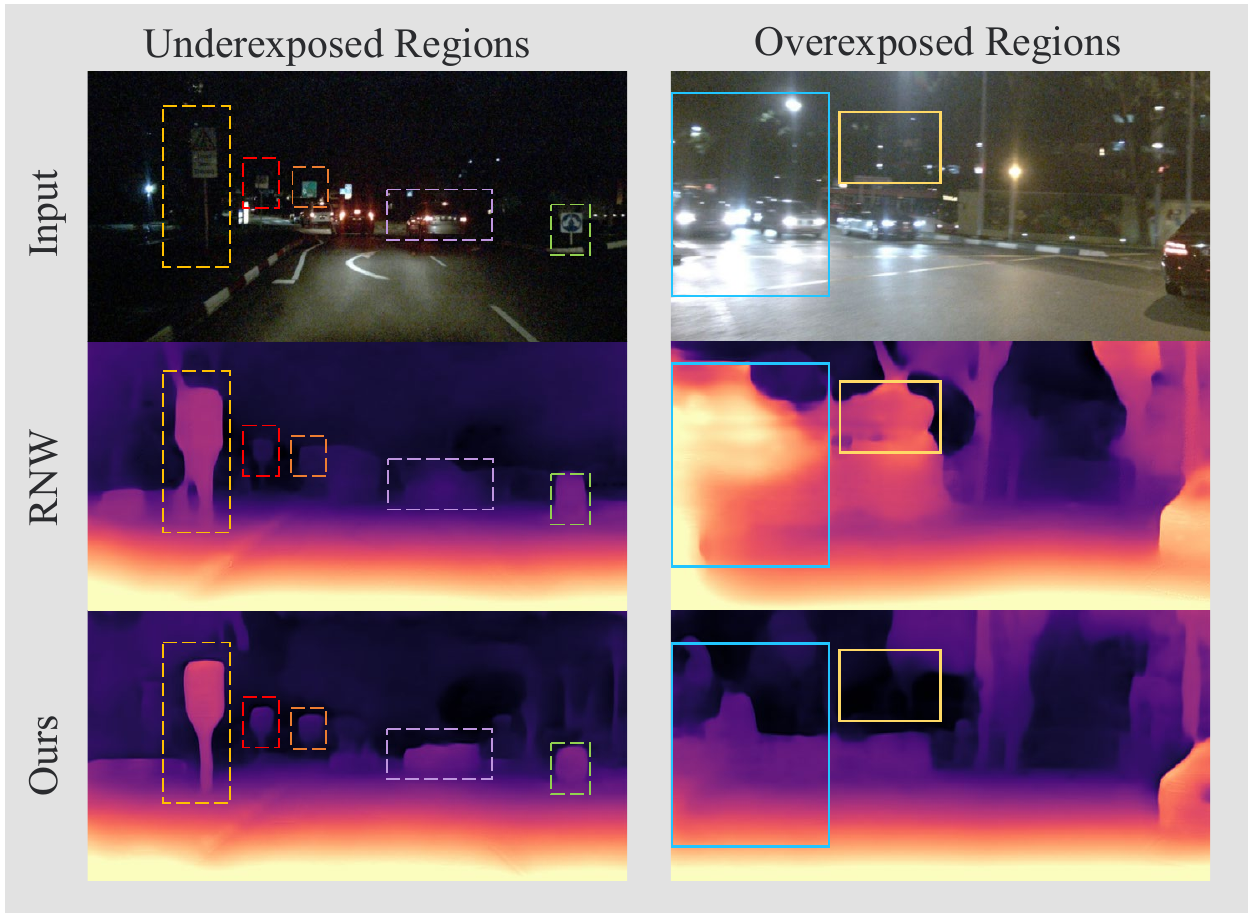}
  \caption{The first row is the input image of two scenes, and the colored dotted and solid boxes mark the objects in underexposed regions (left) and overexposed regions (right). The last two rows are the depth maps predicted by RNW\cite{wang2021regularizing} and our method respectively, indicating the effectiveness of image enhancement and uncertain pixel masking in underexposed and overexposed regions.}
  \label{fig:teaser}
\end{figure}

Pointcloud-based sensing algorithms are of great significance to computer vision society\cite{bu2019pedestrian,ryu2021scanline,vodisch2022end,bai2022faster,mccraith2022lifting, huang2021spatio, chen2020boost, chen2022pq, han2022fast, tian2022vibus, zhang2018efficient, jin2022language}. However, they are widely considered too expensive for autonomous vehicles. In this regard, image-based depth estimation has drawn a lot of attention from the robotics community \cite{jing2022depth, licuaret2022ufo, choi2022selftune, patil2022improving, boulahbal2022instance, chen2018multi} due to low hardware cost. Amongst learning-based depth estimation methods, self-supervised formulations using image sequences \cite{Fu},\cite{Kuznietsov} are quite appealing as they do not require paired RGB-D data and open up the opportunity for online adaptation \cite{li2021generalizing}. And with the efforts of \cite{godard2019digging,guizilini20203d,8793512,8793621}, the performance of self-supervised depth estimation on KITTI \cite{KITTI}, Cityscapes \cite{cordts2016cityscapes}, and DDAD \cite{guizilini20203d} datasets is already comparable to supervised methods. However, these studies all focus on daytime image sequences where inputs are well-lit and the photometric consistency assumption generally holds. Self-driving vehicles need to run robustly during nighttime and unfortunately the photometric consistency assumption hardly holds in this challenging scenario. 

A natural idea is to use nighttime image enhancement techniques to improve the quality of input images. But supervised nighttime image enhancers are intrinsically restricted by dataset bias as existing paired day/night datasets focus on indoor scenes and building these kinds of paired datasets for dynamic road scenes is nearly impossible. To this end, we propose the first learning framework that learns an enhancer and a single-view depth estimator both in a self-supervised manner, as shown in Fig. \ref{fig:main}. Since these two modules can collaboratively work towards a common goal without any ground truth, our method significantly outperforms the previous SOTA method RNW \cite{wang2021regularizing} as shown in Fig. \ref{fig:teaser}.

Delving into this framework, we identify an interesting but overlooked fact: nighttime images suffer not only from underexposed regions but also  from overexposed regions (referred to as \emph{unexpected regions} as a whole later). Both cause some detailed information loss and prevent the model from estimating accurate depth through local contextual cues. Moreover, the overexposed regions are often associated with the movement of cars (e.g., car light), which also violates photometric consistency. In the spirit of influential auto-masking techniques in the self-supervised depth estimation literature \cite{godard2019digging}\cite{ranjan2019competitive}, we strive to softly mask out unexpected regions.

We observe that the mid-product of image enhancement -- enhancement ratio (or say the illumination component) could provide hints to find the unexpected regions, \emph{i.e.}, underexposed areas need a higher ratio and vice versa. This observation motivates us to design an uncertainty map built on the ratio to suppress unexpected areas. After that, we bridge depth estimation and image enhancement tightly, using a bridge-shaped model (Fig. ~\ref{fig:mask}) for soft masking. Apart from that, we also introduce a pre-trained denoising module to increase the image signal-to-noise ratio.

Furthermore, we find that the existing night driving datasets have only sparse ground truth due to the limit of LiDAR data which cannot cover all areas of interest during evaluation. 
Following the idea that transferring the knowledge in the simulation environment to the real world, we resort to CARLA \cite{dosovitskiy2017carla}, the simulator for autonomous driving research.
However, huge domain gap between the rendered images and the real-world images makes it not straightforward to use the simulated data directly.
Thus, we propose CARLA-EPE, a photo-realistically enhanced nighttime dataset based upon CARLA.
We leverage the state-of-the-art photorealism enhancement network EPE \cite{richter2022enhancing} to transfer the style of the rendered images to a photorealistic one, resulting in a photorealistic nighttime dataset CARLA-EPE with dense ground truth of depth.
From the experiment results, the task in our new dataset is more challenging than others, which brings meaningful new challenges to the field.

In brief, our contributions can be summed up in four-fold:
\begin{itemize}
\item[$\bullet$] We propose the first method that jointly learns nighttime image enhancement and depth estimation without using any paired ground truth.
\item[$\bullet$] We identify that the illumination component in the self-supervised nighttime image enhancer can be used to identify unexpected regions and propose a bridge-shape model for soft auto-masking.
\item[$\bullet$] We contribute a novel photorealistically enhanced nighttime dataset with dense depth ground truth.
\item[$\bullet$] We achieve SOTA performance on public benchmarks and release our codes.
\end{itemize}

\section{Related Works}

\subsection{Self-supervised Monocular Depth Estimation} 

Various methods have studied self-supervised depth estimation in computer vision owing to its less supervision demanded. SfMLearner\cite{zhou2017unsupervised} is the first work to address this problem by jointly learning the depth and relative pose between two adjacent frames to reconstruct the target frame with photometric consistency loss. Since the photometric constraint is easily affected by the blur, occlusion, and moving objects, several works proposed effective approaches, such as optical flow\cite{yin2018geonet},  instance segmentation\cite{casser2019depth}, stationary pixel mask\cite{godard2019digging}, point cloud consistency\cite{Mahjourian}, and packing network\cite{guizilini20203d}.
However, these methods are designed for daytime depth estimation and fail to work in nighttime scenarios. The biggest challenge of nighttime depth estimation is that the assumption of temporal illumination consistency is invalid due to the low-light and non-uniform illuminations. Recently, a few works have explored it. ADDS \cite{liu2021self} proposed a domain-separated network to extract illumination and texture features for both day-time and nighttime depth estimation networks. Different from ADDS, RNW \cite{wang2021regularizing} trained on unpaired data by (1) leveraging a GAN-based method to adapt the daytime estimation network to apply for the nighttime data, (2) employing depth distribution from daytime to regularize the nighttime network, and (3) using a HE-based offline image enhancement module to deal with the low-light regions. Despite their better results, these models still suffer from underexposed and overexposed regions.

\subsection{Low-light Image Enhancement}

Low-light image enhancement aims to restore details in low visibility regions. Histogram equalization (HE) \cite{pizer1987adaptive}, including its variants CLHE\cite{pizer1987adaptive}, is a classical image enhancement method that increases the global contrast of images.
However, the HE-based approaches are based upon global distribution rather than local context, so the useable signal may be reduced while background noise contrast is increased. Retinex model-based methods \cite{land1977retinex, jobson1997multiscale, jobson1997properties, guo2016lime} are another alternative ways for image enhancement. It assumes the low-light image can be decomposed into illumination and reflectance. Moreover, the reflectance is regarded as the result of enhancement. To better leverage the great power of deep learning, network-based Retinex method \cite{wei2018deep,zhang2021beyond, liu2021retinex} combined CNNs and Retinex theory to pursue better accuracy and robustness. More recently, SCI \cite{ma2022toward} built a cascaded illumination learning process and developed a self-supervised framework. Nevertheless, their generalization performance in challenging driving scenarios is not satisfactory. In addition, a pre-trained image enhancement network is not necessarily suitable for nighttime depth estimation.
\section{Method}\label{sec:method}
\begin{figure*}
  \centering
  \includegraphics[width=0.95\textwidth]{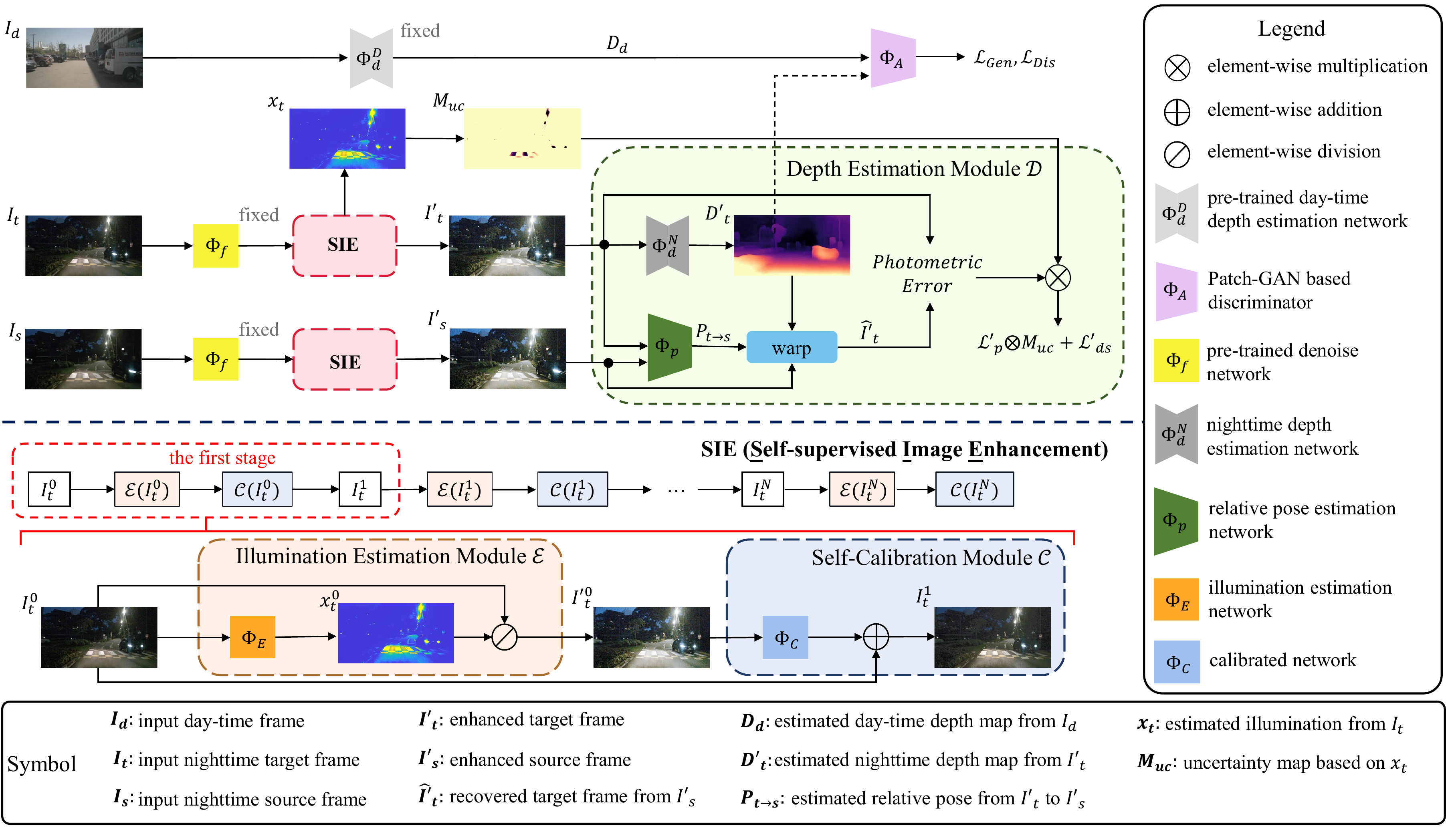}
  \caption{The architecture of our proposed framework (see section \ref{sec:method} for details)  
  }
  \vspace{-0.2cm} 
  \label{fig:main}
\end{figure*}

\subsection{Self-supervised Nighttime Depth Estimation}
Given a single image $I_t \in \mathbb{R}^{H\times W\times 3}$, the goal of learning-based depth estimation is to predict a depth map $D_t\in\mathbb{R}^{H\times W}$ by a trainable network $\Phi_d: \mathbb{R}^{H\times W\times 3} \rightarrow\mathbb{R}^{H\times W}$.
To achieve self-supervision, the key idea is to reconstruct the target frame $I_t$ from the source one $I_s$ according to the geometry constraint. To be specific, given a known camera intrinsic matrix $K$, a predicted depth map $D_t$ and relative pose $P_{t \to s}\in\mathbb{R}^{4\times4}$ between source and target frames via a trainable network $\Phi_{p}: \mathbb{R}^{H\times W\times 6} \rightarrow\mathbb{R}^{4\times4}$, each point $p_t$ in $I_t$ can be projected onto the source view $p_s$ in $I_s$ by
\begin{equation}\label{eq:synthesis}
    p_s \sim KP_{t \to s}D_t(p_t)K^{-1}p_t{,}
\end{equation}
where $\sim$ represents the homogeneous equivalence. With Eq. \ref{eq:synthesis}, we can recover the target frame $\hat I_t$ from $I_s$ by:
\begin{equation}
    {\hat I_t}=I_{s}\bigg {\langle} {\rm proj}(D_t, P_{t \to s}, K) \bigg \rangle {,}
\end{equation}
where $\big < \cdot \big >$ is a differentiable bilinear interpolation proposed by \cite{STN} and $\rm proj(\cdot,\cdot,\cdot)$ is the projection operation in Eq. \ref{eq:synthesis}.

The training signal is the photometric error between the target frame $I_t$ and the reconstructed frame $\hat I_t$. 
Following \cite{godard2019digging}, we combine $\ell_1$ and SSIM losses as the photometric loss $L_p$, which is defined as: 
\begin{equation}\label{eq:pe_loss}
   \mathcal{L}_p(I_t, \hat I_t) =\alpha \frac{(1-\rm SSIM(I_t,\, \hat I_t))}{2}+(1-\alpha) \vert I_t-,\hat I_t \vert_1 {,}
\end{equation}
where we set $\alpha = 0.85$ for all experiments as in \cite{godard2019digging}.
Additionally, as stated in \cite{godard2019digging}, minimizing Eq. \ref{eq:pe_loss} can only enforce a necessary but not sufficient condition. Therefore, we follow \cite{godard2019digging} to avoid depth ambiguity by enforcing smoothness of the predicted depth map, \emph{i.e.} 
\begin{equation}\label{eq:smooth}
    \mathcal{L}_{ds}(D_t, I_t)= |\partial_x D_t|e^{-|\partial_x I_t|}+ |\partial_y D_t|e^{-|\partial_y I_t|} {.}
\end{equation}

Due to the poor quality of nighttime images, training the learning system with Eq. \ref{eq:pe_loss} can only provide noisy gradients. To alleviate that, we follow \cite{wang2021regularizing} to introduce a pre-trained day-time depth estimation model as prior to direct the nighttime model training via an adversarial manner. As shown in Fig. \ref{fig:main}, we build a nighttime depth estimation network $\Phi^N_{d}$ as a generator, aiming to make its prediction $D_t$ indistinguishable with $D_d$ which is the output of a pre-trained and fixed day-time depth estimation network $\Phi^D_{d}$. A Patch-GAN-based discriminator $\Phi_{A}$ is a trainable network to distinguish $D_t$ and $D_d$. $\Phi^{N}_{d}$ and $\Phi_{A}$ are trained by minimizing the GAN-based loss functions, which are formulated as:
\begin{equation}
    \mathcal{L}_{\rm{Dis}} = \frac{1}{2|I_d|}\sum_{D_d}(\Phi_A(D_d)-1)^{2} + \frac{1}{2|I_t|}\sum_{D_t}(\Phi_A(D_t))^{2} {,}
\end{equation}
\vspace{-0.2cm} 
\begin{equation}
    \mathcal{L}_{\rm{Gen}} = \frac{1}{2|I_t|}\sum_{D_t}(\Phi_A(D_t)-1)^{2} {,}
\end{equation}
where $|I_d|$ and $|I_t|$ are the number of day-time and nighttime training images, $D_d=\Phi^D_d(I_d)$ and $D_t=\Phi^N_d(I_t)$. 

\subsection{Joint Training Framework}
As discussed before, nighttime image enhancement could improve the quality of input images to help depth estimation. But supervised nighttime image enhancers are intrinsically restricted by dataset bias.
Hence, we propose a framework to jointly train depth estimation and image enhancement (SIE) in a self-supervised manner, as shown in Fig. \ref{fig:main}. 

According to the Retinex theory\cite{land1977retinex}, given a low-light image $I_t$, the enhanced image can be obtained by $I'_t = I_t \oslash x$, where $x$ is the illumination map which is the most vital part of image enhancement. An inaccurate illumination estimation may bring an over-enhancement result. In order to improve performance stability and decrease computational burden, we follow the stage-wise illumination estimation with a self-calibrated module structure from SCI \cite{ma2022toward}, as shown in the bottom of Fig. \ref{fig:main}.
The enhancement process is formulated by
\begin{equation}
    I_t^{n+1} = \mathscr{C}(\mathscr{E}(I_t^{n})) {,}
\end{equation}
where $n$ ($0<n<3$) is the stage, $\mathscr{E}$ and $\mathscr{C}$ represent illumination estimation and calibration module respectively. 
For stage $n$, $\mathscr{E}$ and $\mathscr{C}$ are implemented by
\begin{equation}
\mathscr{E}(I_t^{n}) = \left\{  
             \begin{array}{lr}  
             x_t^{n} = \Phi_E (I_t^{n}) &  \\  
             I_{t}^{'n} = I_{t}^{n}  \oslash x^{n}_{t},\\  
             \end{array}  
\right.
\end{equation}
\vspace{-0.2cm} 
\begin{equation}
\mathscr{C}(I_t^{'n}) = \left\{  
             \begin{array}{lr}  
             {\rm {res}}^{n}_{t} = {\Phi}_C (I'^{n}_t{}) &  \\  
             I_t^{n+1} = I_t^{n} \oplus {\rm {res}}_t^{n},\\
             \end{array}  
\right.
\end{equation}
where $\Phi_E$ and $\Phi_C$ are the trainable networks to estimate illumination $x^n_t$ and generate calibrated residual map $\rm {res}_t^{n}$ respectively. $\Phi_E$ and $\Phi_C$ share the same parameters in each stage. The calibration module re-generates a pseudo nighttime image so that SIE can be applied in several stages and empirically calibration brings faster convergence and better enhancement.
The enhancement loss contains fidelity and smoothness loss, formulated as
\begin{equation}\label{eq:lf}
    \mathcal{L}_f = \frac{1}{|I_t|} \sum_{I_t} \sum_{n} \vert \vert x_t^{n} - I_t^{n} \vert \vert_2 {,}
\end{equation}
\begin{equation}
    \mathcal{L}_{es} = \frac{1}{|I_t|} \sum_{I_t} \sum_{n} \sum_{i} {\sum_{j \in \mathscr{W}(i)} {\kappa_{i,j}\vert x_t^{n}(i) - x_t^{n}(j)  \vert}} {,}
\end{equation}
where $\kappa_{i,j} $ is weight of a gaussian kernel, $\mathscr{W}(i)$ is a window centered at $i$ with $ 5 {\times} 5 $ adjacent pixels, and $x(i)$ means the pixel value of $x$ at $i$. 
The insight behind $L_f$ is that for nighttime images, the illumination component is largely similar to the input image. Meanwhile $L_{es}$ is a consistency regularization loss.

In Fig. \ref{fig:main}, for joint training, the enhanced result, \emph{i.e.}, the first stage output of SIE, is the input of $\Phi_{d}$ and $\Phi_{p}$. Besides, during the warping (Eq. \ref{eq:synthesis}) and loss computing (Eq. \ref{eq:pe_loss}, \ref{eq:smooth}), the target frame $I_t$ and the recovered frame $\hat I_t$ are replaced by the enhanced frame $I'_t$ and $\hat I'_t$ respectively. We denote the modified losses in Eq. \ref{eq:pe_loss}, \ref{eq:smooth} as $\mathcal{L'}_p(I'_t, \hat I'_t)$ and $\mathcal{L'}_{ds}(D'_t, I'_t)$.

\subsection{Statistics-Based Pixel-wise Uncertainty Mask}
Nighttime images often contain overexposed or underexposed regions where important details would be lost, as shown in Fig. \ref{fig:teaser}. It extremely breaks the model from predicting accurate depth via local contextual cues. Moreover, the overexposed regions are often associated with the movement of cars (e.g., car light), which also violates the illumination consistency in Eq. \ref{eq:pe_loss}. Therefore, we need to design a certain mechanism to bypass such regions. The success of Monodepth2 \cite{godard2019digging} demonstrates that the masking strategy is a simple and effective way to filter out regions that do not meet the assumption of photometric consistency. Inspired by them, we strive to softly mask out unexpected regions. 

\begin{figure}
  \centering
  \includegraphics[width=0.49\textwidth]{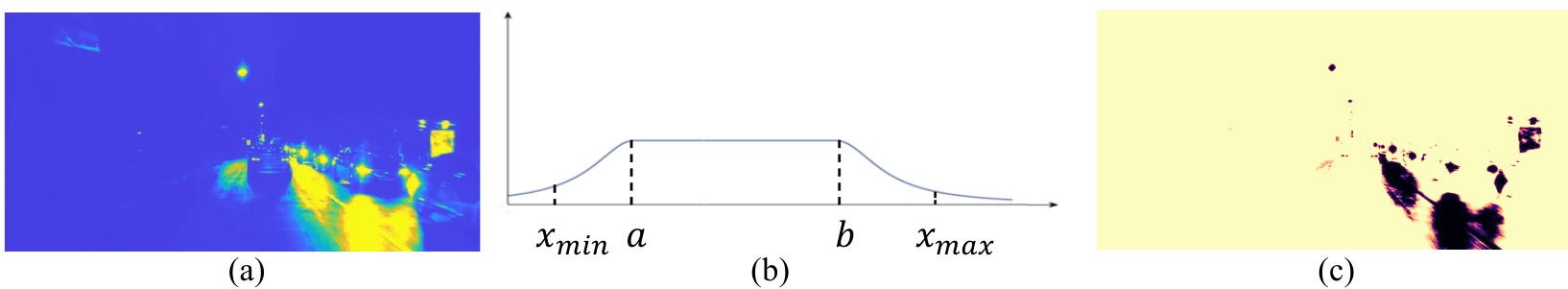}
  \caption{This figure shows the principle of $M_{uc}$. (a) is the illumination map $x_t$. (b) shows the $M_{uc}$ function, which could softly mask out unexpected regions. (c) is the visualization of uncertainty mask.}
  \vspace{-0.4cm} 
  \label{fig:mask}
\end{figure}

Recall that we have used an image enhancement module, SIE. It can predict an illumination map  ${\rm x}_t$ to determine the enhancement ratio of the color of each pixel. As shown in Fig. \ref{fig:main}, the ratio tends to be large in the underexposed regions and small in the overexposed regions. If we use this to weigh the importance of each pixel in photometric loss, the unexpected regions, as mentioned above, are more likely to be filtered. 
Specifically, we define an uncertainty map $M_{uc} \in \mathbb{R}^{H\times W}$, which gives low confidence in the unexpected regions and high confidence in the reasonable regions. $M_{uc}$ is built upon $x_t$ and formulated as
\begin{equation}
M_{uc} = \left\{  
         \begin{array}{lr}  
         {\frac{1}{1 + p^2(x - a)^2}} , & x_{min} \leq x \le a  \\  
         1, & a \leq x \leq b \\
         {\frac{1}{1 + q^2(x - a)^2}} , & b \le x \leq x_{max} \\
         \end{array}  
\right.
\end{equation}
where $a$ and $b$ are the statistics-based bound of illumination to filter reasonable areas, $p$ and $q$ are the attenuation coefficients. Intuitively, this function looks like a bridge as shown in Fig.\ref{fig:mask}. Suggested values of $a$, $b$, $p$ and $q$ can be found in the code release.

\subsection{Image Denoising}
Image denoising is another useful component in the nighttime depth estimation since the low-light images captured by the sensor in night scenes contain much more noise. Especially after the low-light image is enhanced, the noise is inevitably amplified. It could affect the performance of the training signal as it further breaks the illumination consistency between adjacent frames. We use a network AP-BSN\cite{lee2022ap} with the model pre-trained on the SIDD\cite{abdelhamed2018high} dataset. In order to decrease the training burden, the weights of the denoising network AP-BSN are fixed during training.
\subsection{Full Pipeline}
In summary, for SIE module, the total loss ${\mathcal{L}}_{\rm{SIE}}$ is formulated by
\begin{equation}
    {\mathcal{L}}_{\rm{SIE}} = \beta \mathcal{L}_f + \gamma \mathcal{L}_{es} {,}
\end{equation}
and for depth estimation module, the total loss ${\mathcal{L}}_{\rm{DE}}$ is formulated by
\begin{equation}
    {\mathcal{L}}_{\rm{DE}} = \lambda \mathcal{L}'_p \otimes M_{uc} + \mu \mathcal{L}'_{ds} {.}
\end{equation}
The total loss of the whole pipeline is defined as
\begin{equation}
    \mathcal{L}_{\rm{total}} = \eta {\mathcal{L}}_{\rm{SIE}} + \zeta {\mathcal{L}}_{\rm{DE}} + \xi \mathcal{L}_{\rm{Gen}} + \rho \mathcal{L}_{\rm{Dis}} {.}
\end{equation}
The $\beta, \gamma, \lambda, \mu, \eta, \zeta, \xi$, and $\rho$ are hyper-parameters.


\section{Experiments}
\subsection{Dataset}
\begin{figure*}
  \centering
  \includegraphics[width=0.9\textwidth]{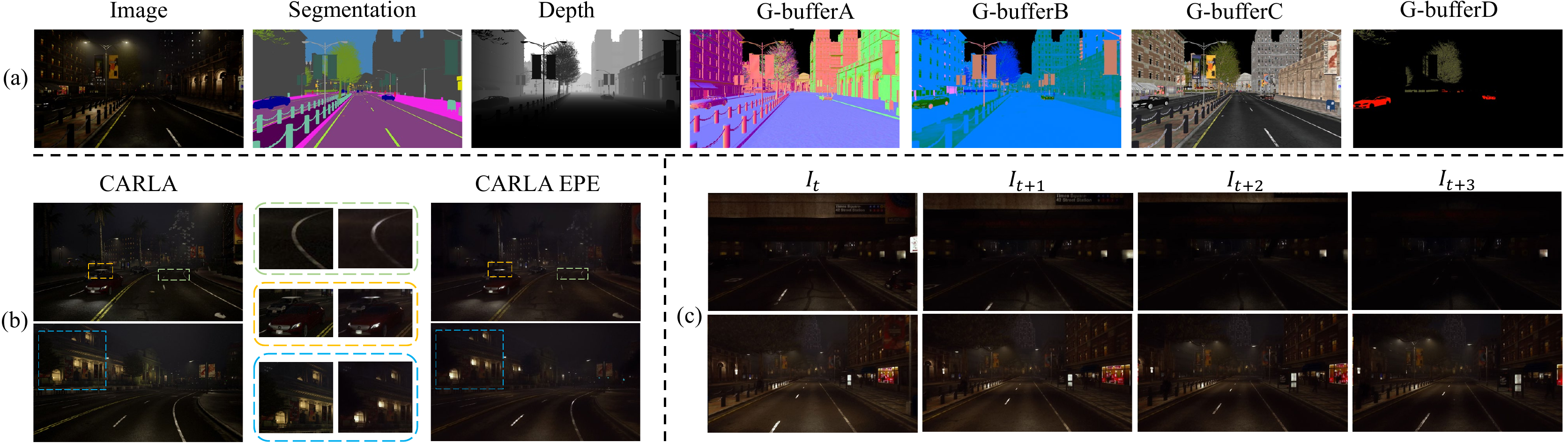}
  \caption{G-buffers and CARLA-EPE Dataset. (a) shows G-buffers of the rendered image from CARLA, representing geometry (Depth, G-bufferA), materials (G-bufferB, G-bufferC) and motion (G-bufferD). (b) is the comparison of the original CARLA image and the corresponding CARLA-EPE image. The dashed box clearly demonstrates the enhancement of details. The orange and green dotted boxes in the middle show the details of the specular highlights from the car roof and lane lines under the street lights, and the blue dotted boxes show the more realistic lighting in the night scene. (c) contains two continuous sequences in CARLA-EPE, which exhibits drastic lighting changes between adjacent frames.}
  \vspace{-0.3cm} 
  \label{fig:EPE}
\end{figure*}

\textbf{CARLA-EPE.} Existing datasets for road scene depth estimation exploit LiDAR to acquire the groundtruth depth, which can only generate sparse depth maps and require a higher cost. The sparse ground truth may not reveal the overall performance of depth estimation methods. 
Although the RGB image and the corresponding dense depth map can be easily collected in the simulator (e.g., CARLA \cite{dosovitskiy2017carla}), the large domain gap between the simulated and the real image dramatically affects the application of the trained model in real scenes.
Hence, we propose a nighttime depth estimation dataset based on CARLA and the enhancing photorealism enhancement (EPE) network \cite{richter2022enhancing}, which can provide dense depth ground truth and photo-realistic images. The pipeline of the dataset generation is as follows. 
Inspired from \cite{richter2016playing} and \cite{richter2017playing}, we first capture rendered images as well as intermediate rendering buffers (G-buffers), which contains geometry, materials and motion information shown in Fig. \ref{fig:EPE} (a) from CARLA simulator.
Then we find matches between rendered dataset and the DarkZurich dataset \cite{sakaridis2020map} taken from the real-world with semantic label predicted by DANNet\cite{wu2021dannet}.
After that, we train the EPE \cite{richter2022enhancing} network to transfer the render images to realistic styles of DarkZurich.
As for depth, we use the depth channel, one of the G-buffers we extract.
We finally generate 12,000 pairs of nighttime images and dense depth maps. 
Fig. \ref{fig:EPE} (b) shows some comparisons of original and enhanced images. It shows that specular highlights on cars and lane lines to cars and lane lines under street lights are visible and building lighting is more realistic at night. This significantly increases the realism of the rendered image.
Fig. \ref{fig:EPE} (c) shows two image sequences whose brightness varies greatly between adjacent frames due to changes in lighting, which is similar to real night scenes.

\textbf{nuScenes\cite{caesar2020nuscenes}.} nuScenes is a large autonomous driving dataset comprising 1000 video clips collected in diverse road scenes and weather conditions. These scenes are pretty challenging, with a fair amount of unexpected regions. 

\textbf{RobotCar\cite{maddern20171}.} RobotCar is a large-scale autonomous driving dataset including driving videos captured on a consistent route during various weather conditions, traffic conditions, and times of day and night. 

\begin{figure*}
  \centering
  \includegraphics[width=0.8\textwidth]{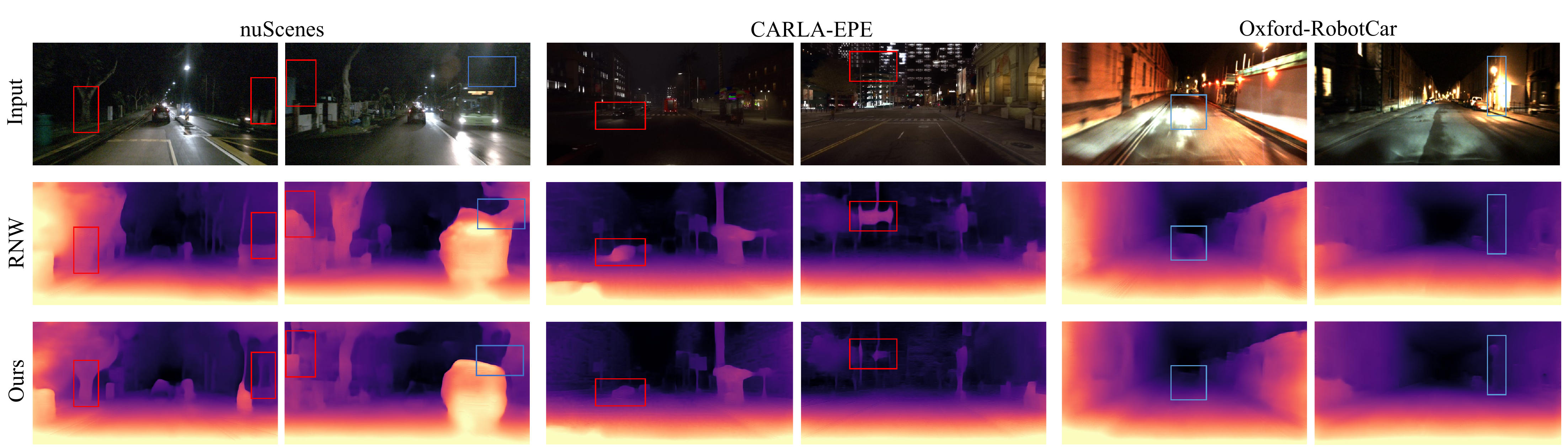}
  \caption{Qualitative results on nuScenes, CARLA-EPE, and RobotCar datasets.}
  \label{fig:q1}
  \vspace{-0.2cm} 
\end{figure*}

\subsection{Implementation Details}
During test, we only evaluate predictions where the groundtruth depth is within 60 meters (m) in RobotCar and nuScenes, and 40 m in CARLA-EPE. For hyper-parameters in training process, please check the code release.



\begin{table}[]
\begin{center}
\caption{Comparison with other methods. Lower is better for error and higher is better for accuracy.}
\label{tab:Quantitative}
\resizebox{\columnwidth}{!}{
\begin{tabular}{ccccccccc}
\cline{3-9}
 & \multicolumn{1}{c|}{} & \multicolumn{4}{c|}{\textbf{Error}} & \multicolumn{3}{c}{\textbf{Accuracy}} \\ \hline
\textbf{Method} & \multicolumn{1}{c|}{\textbf{Max Depth}} & \cellcolor[HTML]{C2F5C5}\textbf{Abs Rel} & \cellcolor[HTML]{C2F5C5}\textbf{Sq Rel} & \cellcolor[HTML]{C2F5C5}\textbf{RMSE} & \multicolumn{1}{c|}{\cellcolor[HTML]{C2F5C5}\textbf{RMSE log}} & \cellcolor[HTML]{F7E196}\textbf{a1} & \cellcolor[HTML]{F7E196}\textbf{a2} & \cellcolor[HTML]{F7E196}\textbf{a3} \\ \hline
\multicolumn{9}{c}{\cellcolor[HTML]{F2BCF7}\textbf{nuScenes}} \\ \hline
\textbf{MonoDepth2\cite{godard2019digging}} & \multicolumn{1}{c|}{60 m} & 1.185 & 42.306 & 21.613 & \multicolumn{1}{c|}{1.567} & 0.184 & 0.360 & 0.504 \\ \hline
\textbf{RNW\cite{wang2021regularizing}} & \multicolumn{1}{c|}{60 m} & 0.326 & 3.999 & 9.932 & \multicolumn{1}{c|}{0.417} & 0.492 & 0.765 & 0.870 \\ \hline
\textbf{Ours} & \multicolumn{1}{c|}{60 m} & \textbf{0.292} & \textbf{3.363} & \textbf{9.120} & \multicolumn{1}{c|}{\textbf{0.390}} & \textbf{0.572} & \textbf{0.805} & \textbf{0.908} \\ \hline
\multicolumn{9}{c}{\cellcolor[HTML]{B183E0}\textbf{RobotCar}} \\ \hline
\textbf{MonoDepth2\cite{godard2019digging}} & \multicolumn{1}{c|}{60 m} & 0.580 & 21.446 & 12.771 & \multicolumn{1}{c|}{0.521} & 0.552 & 0.840 & 0.920 \\ \hline
\textbf{DeFeat-Net\cite{spencer2020defeat}} & \multicolumn{1}{c|}{60 m} & 0.335 & 4.339 & 9.111 & \multicolumn{1}{c|}{0.389} & 0.603 & 0.828 & 0.914 \\ \hline
\textbf{ADFA\cite{vankadari2020unsupervised}} & \multicolumn{1}{c|}{60 m} & 0.233 & 3.783 & 10.089 & \multicolumn{1}{c|}{0.319} & 0.668 & 0.884 & 0.924 \\ \hline
\textbf{ADDS\cite{liu2021self}} & \multicolumn{1}{c|}{60 m} & 0.231 & 2.674 & 8.800 & \multicolumn{1}{c|}{0.268} & 0.620 & 0.892 & 0.956 \\ \hline
\textbf{RNW\cite{wang2021regularizing}} & \multicolumn{1}{c|}{60 m} & 0.185 & 1.894 & 7.319 & \multicolumn{1}{c|}{0.246} & 0.735 & 0.910 & 0.965 \\ \hline
\textbf{Ours} & \multicolumn{1}{c|}{60 m} & \textbf{0.170} & \textbf{1.686} & \textbf{6.797} & \multicolumn{1}{c|}{\textbf{0.234}} & \textbf{0.758} & \textbf{0.923} & \textbf{0.968} \\ \hline
\end{tabular}
}
\vspace{-0.7cm} 
\end{center}
\end{table}

\subsection{Quantitative and Qualitative Results}
Here, we compare our approach with strong baselines on nighttime depth estimation, as shown in Table \ref{tab:Quantitative} and demonstrate more qualitative results in Fig. \ref{fig:q1}.
Table \ref{tab:Quantitative} shows the comparison results with the daytime self-supervised depth estimation method MonoDepth2 \cite{godard2019digging} and other competitive nighttime methods such as DeFeat-Net\cite{spencer2020defeat}, ADFA \cite{vankadari2020unsupervised}, ADDS \cite{liu2021self}, and RNW \cite{wang2021regularizing} on the nuScenes and RobotCar dataset. The metrics in Table \ref{tab:Quantitative} are standard ones used in prior works like \cite{zhou2017unsupervised}. Overall, our method achieves a significant improvement over the other baselines in all metrics on both datasets and sets SOTA. In the following, we choose Abs Rel and $a1$ as the representative metrics to indicate error and accuracy, respectively. On the nuScenes and RobotCar datasets, our method improves the accuracy of RNW by 16.2\% and 3.5\% and reduces the error by 10.4\% and 7.6\%. 
The improvement on the nuScenes dataset is more significant because the overexposed and underexposed regions are prevalent in its scenes, and the images are noisier. This is well aligned with the theoretical expectations of our method, as introduced in the method section.

We also provide the qualitative visualization of the predicted depth maps in Fig. \ref{fig:q1}. The blue boxes show that the RNW suffers severely from overexposure, predicting obviously wrong depth.
The red boxes also show that the RNW cannot estimate the object depth in underexposed regions. Owing to the new framework and soft mask strategy we proposed, our method could predict more reasonable depths in unexpected regions and capture the correct boundary of the small or moving objects in nighttime images (such as lampposts and cars). In addition, our model can reach 25 FPS on a single 2080Ti during inference.
\begin{table}[]
\begin{center}
\caption{Ablation study of each proposed component.}
\label{tab:Ablation1}
\resizebox{\columnwidth}{!}{
\begin{tabular}{cccccc}
\cline{3-6}
 & \multicolumn{1}{l|}{} & \multicolumn{2}{c|}{\textbf{Error}} & \multicolumn{2}{c}{\textbf{Accurancy}} \\ \hline
\multicolumn{1}{c|}{\textbf{Method}} & \multicolumn{1}{c|}{\textbf{Max Depth}} & \cellcolor[HTML]{C2F5C5}\textbf{Abs Rel} & \multicolumn{1}{c|}{\cellcolor[HTML]{C2F5C5}\textbf{RMSE}} & \cellcolor[HTML]{F7E196}\textbf{a1} & \cellcolor[HTML]{F7E196}\textbf{a2} \\ \hline
\multicolumn{6}{c}{\cellcolor[HTML]{F2BCF7}\textbf{nuScenes}} \\ \hline
\multicolumn{1}{c|}{\textbf{Separate Training}} & \multicolumn{1}{c|}{60 m} & 0.325 & \multicolumn{1}{c|}{10.196} & 0.543 & 0.782 \\ \hline
\multicolumn{1}{c|}{\textbf{Joint Training (J.T.)}} & \multicolumn{1}{c|}{60 m} & 0.317 & \multicolumn{1}{c|}{9.779} & 0.546 & 0.786 \\ \hline
\multicolumn{1}{c|}{\textbf{J.T. + $M_h$}} & \multicolumn{1}{c|}{60 m} & 0.314 & \multicolumn{1}{c|}{9.503} & 0.551 & 0.791 \\ \hline
\multicolumn{1}{c|}{\textbf{J.T. + $M_{uc}$}} & \multicolumn{1}{c|}{60 m} & 0.302 & \multicolumn{1}{c|}{9.201} & 0.556 & 0.793 \\ \hline
\multicolumn{1}{l|}{\textbf{J.T. + $M_{uc}$ + Denoise (Full)}} & \multicolumn{1}{c|}{60 m} & \textbf{0.292} & \multicolumn{1}{c|}{\textbf{9.126}} & \textbf{0.572} & \textbf{0.805} \\ \hline
\end{tabular}
}
\vspace{-0.3cm} 
\end{center}
\end{table}

\begin{table}[]
\begin{center}
\caption{Ablation study of the evaluation on sparse and dense depth ground truth.}
\label{tab:Ablation2}
\resizebox{\columnwidth}{!}{
\begin{tabular}{cc|cc|cc}
\cline{3-6}
 & \multicolumn{1}{l|}{} & \multicolumn{2}{c|}{\textbf{Error}} & \multicolumn{2}{c}{\textbf{Accurancy}} \\ \hline
\multicolumn{1}{c|}{\textbf{Method}} & \textbf{Max Depth} & \cellcolor[HTML]{C2F5C5}\textbf{Abs Rel} & \cellcolor[HTML]{C2F5C5}\textbf{RMSE} & \cellcolor[HTML]{F7E196}\textbf{a1} & \cellcolor[HTML]{F7E196}\textbf{a2} \\ \hline
\multicolumn{1}{c|}{\textbf{RNW w. dense}} & 40 m & 1.164 & 9.184 & 0.173 & 0.330 \\ \hline
\multicolumn{1}{c|}{\textbf{Ours w. dense}} & 40 m & 1.121 & 8.992 & 0.174 & 0.336 \\ \hline
\multicolumn{1}{c|}{\textbf{RNW w. sparse}} & 40 m & 0.975 & 8.210 & 0.283 & 0.582 \\ \hline
\multicolumn{1}{c|}{\textbf{Ours w. sparse}} & 40 m & \textbf{0.941} & \textbf{7.987} & \textbf{0.310} & \textbf{0.592} \\ \hline
\end{tabular}
}
\vspace{-0.3cm} 
\end{center}
\end{table}

\subsection{Ablation Study}
Here, we provide ablations on nuScenes dataset for the designs of each component we proposed. 
\begin{figure}
  \centering
  \includegraphics[width=0.4\textwidth]{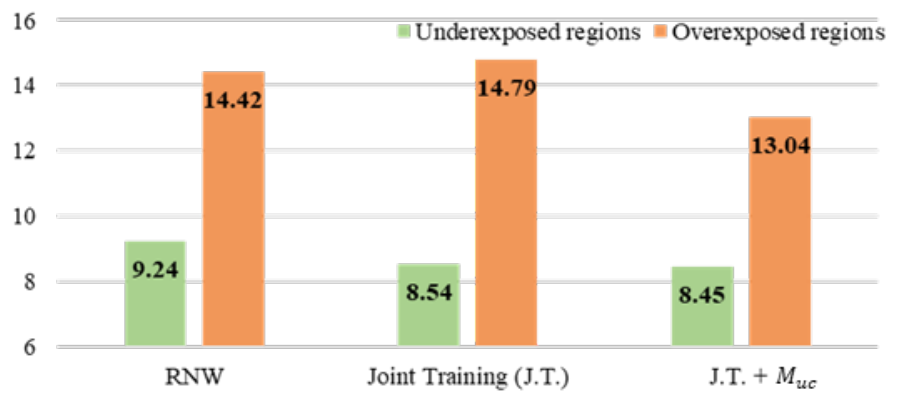}
  \vspace{-0.2cm} 
  \caption{We manually select 63 and 65 images in the test set of nuScenes, containing underexposed and overexposed regions respectively, and then count RMSE of these regions under three settings (RNW, Joint Training (J.T.), and  J.T. + Mask $M_{uc}$).}
  \vspace{-0.5cm} 
  \label{fig:RMSE}
\end{figure}

\textbf{Joint Training Framework}: The result of our joint training framework without masking and denoising is shown in the first row of Table \ref{tab:Ablation1}. We find that it still significantly outperforms the RNW \cite{wang2021regularizing} in Table \ref{tab:Quantitative} and credit it to the new formulation. Since our method makes it possible to learn these two tasks collaboratively towards the final goal, the performance is largely promoted.
The second row demonstrates our method performs better than a baseline that combines a pre-trained SIE with the self-supervised depth estimator, showing the importance of joint self-supervised training.
As expected, Fig. \ref{fig:RMSE} shows that the image enhancement could improve the performance in underexposed regions but not in overexposed regions.

\textbf{Soft Masking}: The fourth row of Table \ref{tab:Ablation1} shows the advantage of the proposed uncertainty masking strategy combined with the joint training framework. Given that the self-supervised enhancement loss $L_f$ (Eq. \ref{eq:lf}) is built upon the insight that input nighttime images are largely proportional to the illumination component. Thus, a natural question is whether it is effective to use the original input image for illumination distribution modelling to suppress unexpected regions as $M_{uc}$ does. So, we build a new baseline called $M_h$ and the result of it shows that (1) it also works, verifying the idea of softly masking out unexpected regions is important and the statistical model well serves its purpose. (2) $M_{uc}$ works better, showing that using the mid-product illumination map is a better choice and the necessity of combining these two tasks because the illumination map is provided by the self-supervised enhancement task. Fig. \ref{fig:RMSE} shows the effectiveness of the masking strategy in overexposed regions. 

\textbf{Denoising}: The last row of Table \ref{tab:Ablation1} indicates the positive impacts of introducing the denoising module, which we credit imporved photometric consistency after denosing.

\begin{figure}
  \centering
  \includegraphics[width=0.48\textwidth]{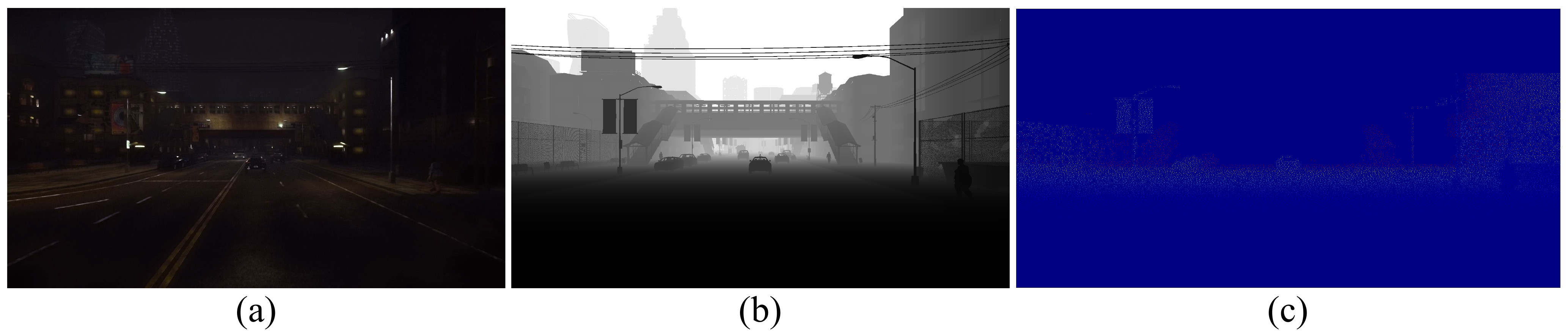}
  \vspace{-0.5cm} 
  \caption{Comparison between sparse and dense ground truth. (a) is the RGB image, (b) is the dense ground truth of the simulator, and (c) is the sparse ground truth from LiDAR. Please zoom in electronically to inspect the sparse depth map.}
  \vspace{-0.5cm} 
  \label{fig:gt}
\end{figure}

\textbf{CARLA-EPE Dataset}: In the first two rows of Table \ref{tab:Ablation2}, our method still outperforms RNW in CARLA-EPE dataset with the dense ground truth, and the dense protocol is more challenging since the results are worse than those on nuScenes dataset with sparse ground truth. Then we sample the dense depth to generate a sparse depth map according to the distribution of the LiDAR data, as shown in Fig. \ref{fig:gt}. As expected, the last two rows of Table \ref{tab:Ablation2} show that the performance is improved in this setting, which indicates the 
limitation of the evaluation with sparse depth maps.

 
\subsection{Conclusion}
In this paper, we propose a self-supervised framework to jointly learn image enhancement and depth estimation. Delving into this framework, we find an effective mid-product of the image enhancement to generate a pixel-wise mask to suppress the overexposed and underexposed regions. Benefiting from these improvements, we set SOTA on public nighttime depth estimation benchmarks. Additionally, we contribute a novel photo-realistically enhanced nighttime dataset with dense depth ground truth.

\subsection{Acknowledgments}
This work was supported by the National Natural Science Foundation of China (NSFC) under Grants No. 62173324 and Tsinghua-Toyota Joint Research Fund (20223930097).

\bibliographystyle{IEEEtran}
\balance
\bibliography{ref}
\end{document}